%% file: example_paper.tex
\documentclass{article}

\usepackage{microtype}
\usepackage{graphicx}
\usepackage{subcaption}
\usepackage{booktabs} 

\usepackage{hyperref}

\usepackage[accepted]{icml2026}

\usepackage{amsmath}
\usepackage{amssymb}
\usepackage{mathtools}
\usepackage{amsthm}

\usepackage{multirow}
\usepackage{graphicx}

\newcommand{\shortcomment}[1]{}
\newcommand{\later}[1]{$\ast\ast\ast$}

\input{math_commands}

\input{my_math_commands}

\usepackage[capitalize,noabbrev]{cleveref}

\theoremstyle{plain}

\usepackage[textsize=tiny]{todonotes}

\icmltitlerunning{Stability Analysis of Sharpness-Aware Minimization}

\begin{document}

\twocolumn[
  \icmltitle{Stability Analysis of Sharpness-Aware Minimization}



  \icmlsetsymbol{equal}{*}

  \begin{icmlauthorlist}
    \icmlauthor{Hoki Kim}{cau}
    \icmlauthor{Jinseong Park}{kias}
    \icmlauthor{Yujin Choi}{unist,ntu}
    \icmlauthor{Jaewook Lee}{snu}
  \end{icmlauthorlist}

  \icmlaffiliation{cau}{Chung-Ang University, South Korea}
  \icmlaffiliation{kias}{Korea Institute for Advanced Study, South Korea}
  \icmlaffiliation{unist}{Ulsan National Institute of Science and Technology (UNIST), South Korea}
  \icmlaffiliation{ntu}{Nanyang Technological University (NTU), Singapore}
  \icmlaffiliation{snu}{Seoul National University, South Korea}

  \icmlcorrespondingauthor{Jaewook Lee}{jaewook@snu.ac.kr}

  \icmlkeywords{Machine Learning, ICML}

  \vskip 0.3in
]



\printAffiliationsAndNotice{}  

\begin{abstract}
Sharpness-aware minimization (SAM) is a training method that seeks to find flat minima in deep learning, resulting in state-of-the-art performance across various domains. Instead of minimizing the loss of the current weights, SAM minimizes the worst-case loss in its neighborhood in the parameter space. In this paper, we investigate the convergence instability of SAM near a saddle point. Using the qualitative theory of dynamical systems, we explain how SAM becomes stuck in the saddle point and theoretically prove that the saddle point can become an attractor under SAM dynamics. Additionally, we show that this convergence instability can also occur in stochastic dynamical systems by establishing the diffusion of SAM. We prove that SAM diffusion is worse than that of vanilla gradient descent in terms of saddle point escape. Finally, we demonstrate that often overlooked training tricks, momentum and batch-size, might be important to mitigate the convergence instability and achieve high generalization performance. Our theoretical and empirical results are thoroughly verified through experiments on several well-known optimization problems and benchmark tasks.
\end{abstract}

\section{Introduction}\label{sec:introduction}
With a large number of parameters, deep learning models have shown remarkable improvements across a variety of domains. However, these over-parameterized deep learning models may suffer from poor generalization, despite achieving near-zero training loss. 
Recently, researchers have explored the relationship between the geometric characteristics of loss surface and the generalization performance \cite{hochreiter1994simplifying, keskar2017large, li2018visualizing}. Theoretical and empirical studies have demonstrated that the generalization performance is potentially related to the sharpness of loss landscape and that an optimum with a flatter loss landscape can lead to better generalization.

Based on these prior studies, \citet{foret2020sharpness} proposed a new training framework called \textit{sharpness-aware minimization} (SAM) that substantially improves generalization performance in various tasks \cite{foret2020sharpness, zhuang2021surrogate, chen2022when}. The key idea of SAM is to minimize the maximum loss near its neighborhood in the weight space instead of minimizing the loss of current weight. SAM demonstrates its ability to find an optimum with a flatter loss landscape, resulting in improved generalization performance. Due to its simplicity and ease of implementation, SAM has been widely adopted across various applications, consistently improving generalization performance \cite{foret2020sharpness, chen2021vision}.

\begin{figure}[t!]
    \centering
    \includegraphics[width=0.85\linewidth]{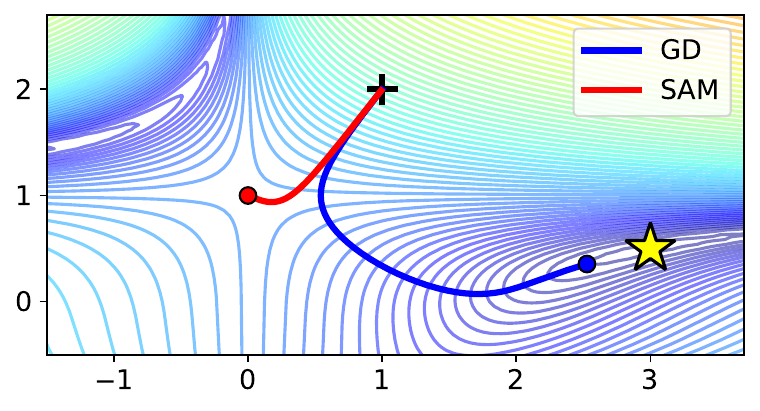} 
    \caption{The optimization has begun at the point indicated by the plus sign, and the global minimum is indicated by the yellow star. SAM appears to be stuck at the saddle point, rather than converging to the global minimum.}
    \label{fig:intro}
\end{figure}

In this paper, we investigate the convergence instability of SAM dynamics near a saddle point. As shown in \Figref{fig:intro}, the Beale function, a widely used optimization problem, is optimized using both vanilla gradient descent (GD) and SAM. GD reaches the global minimum (yellow star), but SAM becomes trapped at the saddle point. As deep learning objectives are highly non-linear and have many local minima and saddle points \cite{du2017gradient, kleinberg2018alternative}, we provide an intuitive explanation of the convergence instability of SAM near saddle points.

The main contributions can be summarized as follows:
\begin{itemize}
    \item We investigate convergence instability of SAM by theoretically demonstrating the difficulty of escaping saddle points under SAM dynamics.

    \item We extend our analysis to stochastic dynamics with diffusion and highlight the roles of momentum and batch size in facilitating escape from saddle points.

    \item We empirically validate our theoretical findings across diverse optimization scenarios, ranging from a basic optimization problem to neural network-based benchmark tasks.

\end{itemize}

\section{Background}
Given a loss function $\ell(\cdot)$ and weight parameters $\vw$, the traditional optimization, the so-called empirical risk minimization, tries to minimize the following objective:
\begin{equation}
    \min_\vw \mathcal{L}(\vw; \mathcal{S}) := \frac{1}{n} \sum_{(\vx, \vy) \in \mathcal{S}}\ell(\vw; \vx, \vy),
    \label{eq:erm}
\end{equation}
where $\mathcal{S}=\{\vx_i, \vy_i\}_{i=1}^n$ is a sampled training dataset with $n$ instances. Under the general i.i.d. assumption on data, \eqref{eq:erm} generally yields a feasible solution of $\vw$ on the true distribution of data $\mathcal{D}$,
\begin{equation}
    \min_\vw \mathcal{L}(\vw; \mathcal{D}) := \mathbb{E}_{(\vx, \vy) \sim \mathcal{D}} \left[ \ell(\vw; \vx, \vy) \right].
    \label{eq:erm_true}
\end{equation}
However, in common practice, the i.i.d assumption is often violated by the limitation of training data and model structure. Therefore, a \textit{generalization gap} can be defined as
\begin{equation}
    \mathcal{E}(\vw) = \mathcal{L}(\vw; \mathcal{S}) - \mathcal{L}(\vw; \mathcal{D}).
    \label{eq:truedistribution}
\end{equation}
A low generalization gap indicates a high generalization performance. Therefore, the primary objective of machine learning is to obtain the solution that minimizes both the training loss $\mathcal{L}(\vw; \mathcal{S})$ and the generalization gap $\mathcal{E}(\vw)$. 

To address this issue, recent studies have focused on the \textit{flatness} of the loss landscape as a potential solution \cite{hochreiter1994simplifying, dziugaite2017computing} and provided experimental evidence that a flatter loss landscape tends to have a better generalization performance \cite{jiang2019fantastic}.
A prominent approach in this direction is \textit{sharpness-aware minimization} (SAM) \cite{foret2020sharpness}, which has demonstrated significant improvements in generalization across various tasks and model structures \cite{zhuang2021surrogate, chen2022when}.
SAM aims to minimize the worst-case loss over its parameter neighborhood rather than minimizing the loss of the current parameter. 
Let us denote a vanilla gradient descent algorithm that minimizes the loss of current weight $\vw_t$ at time $t$ as
\begin{equation}
    \vw_{t+1} = \vw_t - \eta \nabla \ell(\vw_t),
    \label{eq:vanila}
\end{equation}
where $\eta$ is a learning rate and $\nabla\ell(\vv)$ is a gradient with respect to its input vector $\vv$ unless specified otherwise. In contrast, SAM minimizes the loss of \textit{perturbed weight} $\vw^p_t$ by using the first-order Taylor approximation:
\begin{align}
    \vw^p_t &= \vw_t + \rho \nabla \ell(\vw_t). \label{eq:wp} \\
    \vw_{t+1} &= \vw_t - \eta \nabla \ell(\vw^p_t),
    \label{eq:sam}
\end{align}
where $\rho$ is a given neighborhood radius. Note that $\rho$ can be normalized with the gradient norm as introduced in \cite{foret2020sharpness}, i.e., $\rho/\|\nabla\ell(\vw)\|$; however, a constant $\rho$ in \eqref{eq:wp} shows similar or higher performance than the normalized version \cite{andriushchenko2022towards}.
Therefore, SAM consistently enforces $\vw$ to have a reduced perturbed loss $\ell(\vw^p)$ within its $\rho$-neighborhood. 

\section{Related Work}
Following the success of SAM, further investigations have been conducted to explore its algorithm and enhance its generalization performance or computational efficiency.
\citet{kwon2021asam} proposed a transformation of parameters to achieve scale-invariant sharpness.
\citet{zhuang2021surrogate} argued that subtracting $\nabla  \ell (\vw)$ from $\nabla  \ell (\vw^p)$ can reduce certain drawbacks in optimization.
More recently, the focus has shifted to understanding the implicit bias of SAM, which includes analyses on implicit bias for different sizes of mini-batch \cite{andriushchenko2022towards}, the balance in weight magnitudes across layers \cite{li2024implicit}, and effectiveness on learning multiple correlated features \cite{springer2024sharpness}.

Beyond performance improvements and the implicit bias effects of SAM, recent studies have also examined its optimization characteristics. Among these works, \cite{kaddour2022when, si2023practical} provide insights into failure modes of SAM from an optimization perspective. In particular, \citet{kaddour2022when} initially observed atypical behavior of SAM in the vicinity of saddle points, which is the primary focus of our study.

\textbf{Comparison to \cite{compagnoni2023sde}}. \citet{compagnoni2023sde} also analyzes SAM near saddle points, but under a fundamentally different set of assumptions. Their results rely on Lipschitz continuity and a small learning rate regime, interpreting noise as a structured term that implicitly smooths the loss. In contrast, we adopt a local dynamical systems perspective and consider perturbation acts as an effective random forcing term that drives trajectories away from the unstable attractor. This perspective is consistent with recent edge-of-stability findings \cite{long2024sharpness}, especially in regimes with large Hessian eigenvalues. Together with these works \cite{long2024sharpness, compagnoni2023sde}, we believe that our results help complete the understanding of the behavior of SAM.

\begin{figure*}[t!]
    \centering
  \subfloat[Lambda Lemma\label{fig:A1}]{%
       \includegraphics[width=0.27\linewidth]{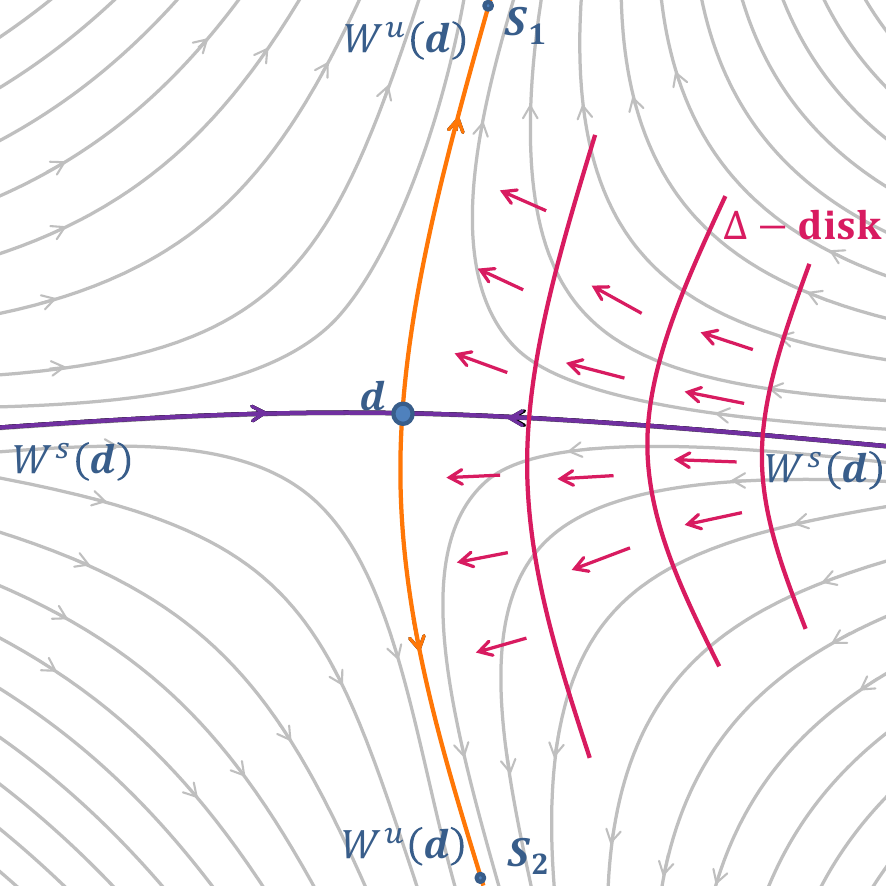}}
    \quad
  \subfloat[Stable and unstable manifold\label{fig:A1-1}]{%
        \includegraphics[width=0.27\linewidth]{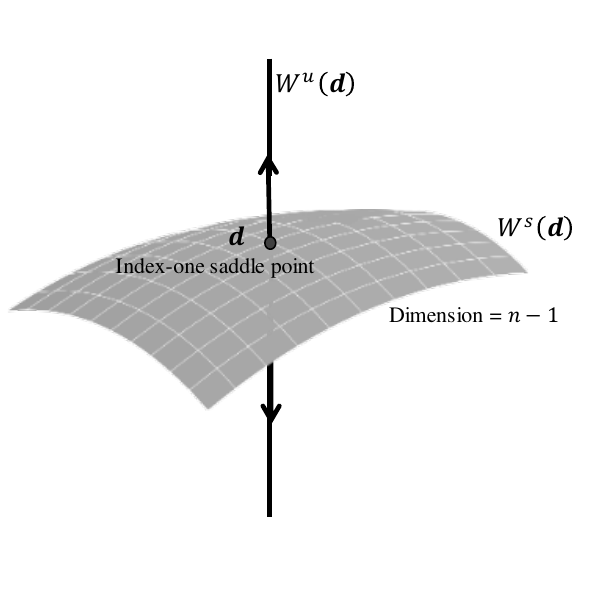}}
    \quad
  \subfloat[Case-I\label{fig:A2}]{%
        \includegraphics[width=0.27\linewidth]{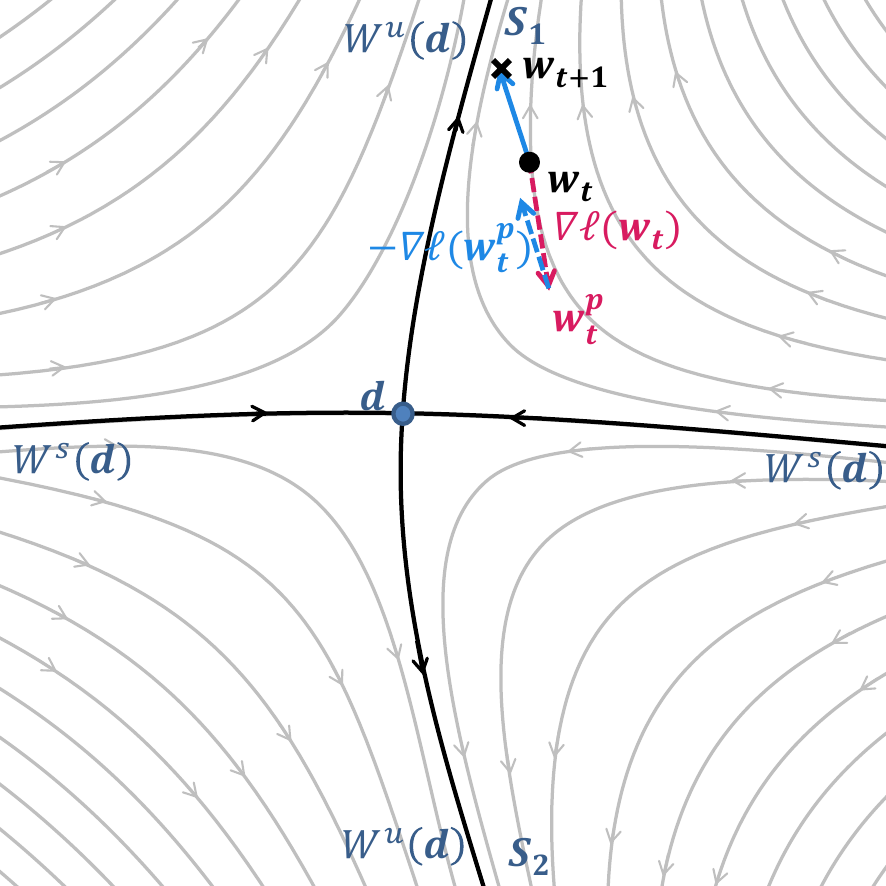}}
    \hfill
  \subfloat[Case-II\label{fig:A3}]{%
        \includegraphics[width=0.27\linewidth]{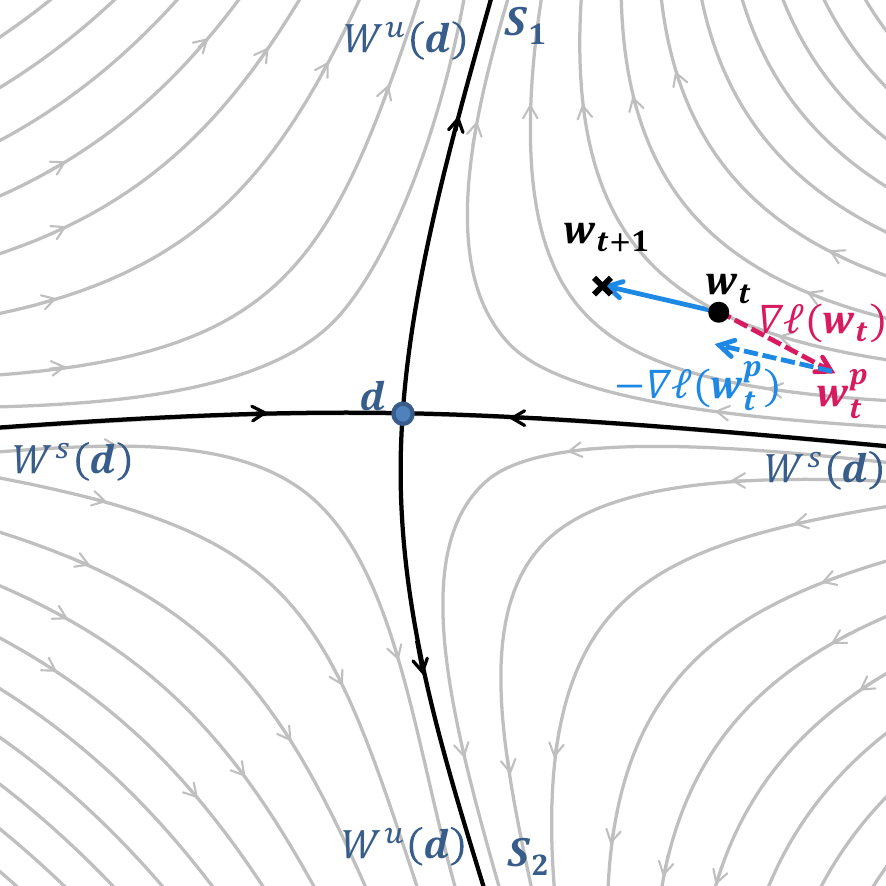}}
    \quad
  \subfloat[Case-III-i\label{fig:A4}]{%
        \includegraphics[width=0.27\linewidth]{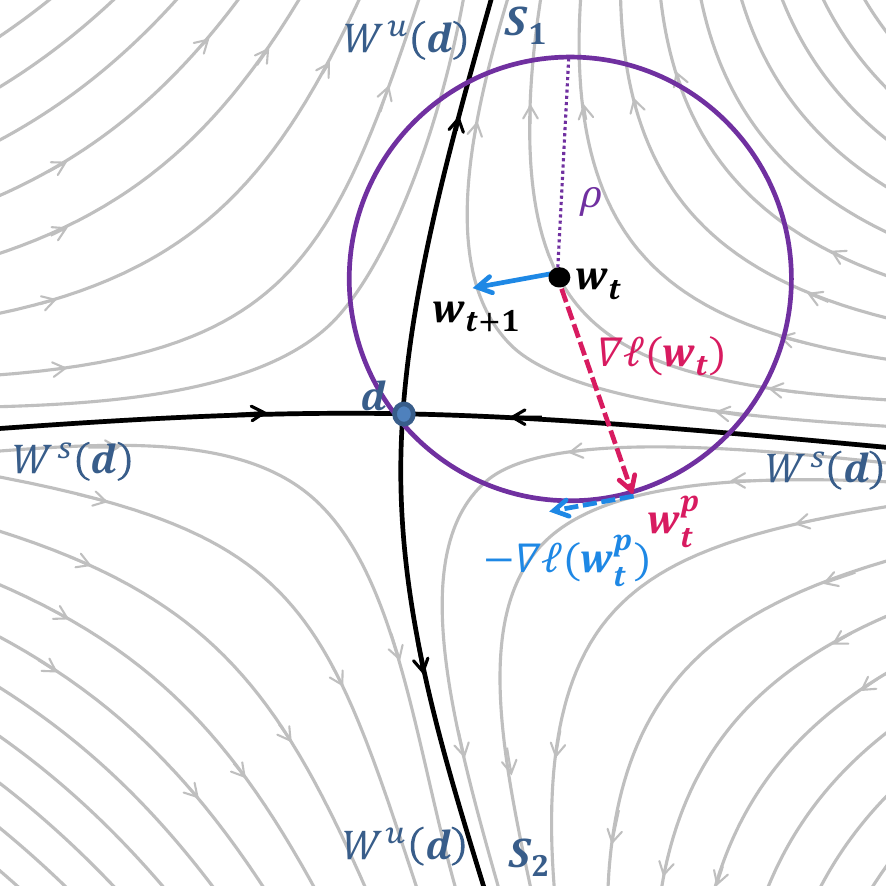}}
    \quad
  \subfloat[Case-III-ii\label{fig:A5}]{%
        \includegraphics[width=0.27\linewidth]{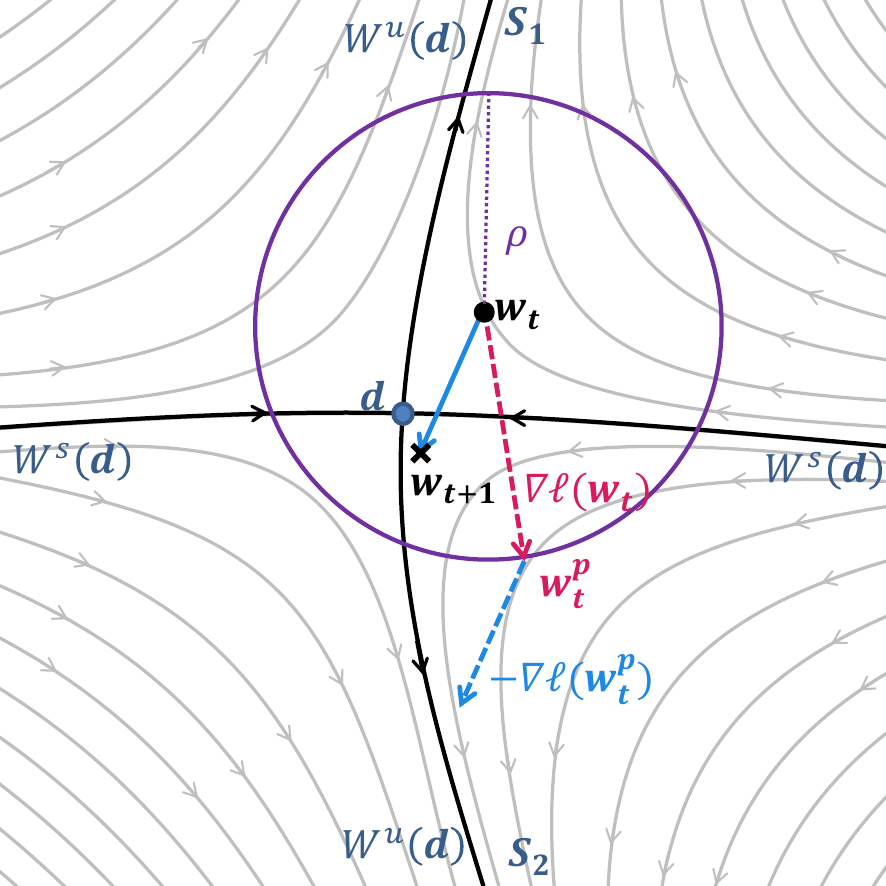}}
  \caption{(a)-(b) Phase portrait of system (\ref{eq:grad}). (c)-(d) Phase portrait of system (\ref{eq:grad}) for Case-I and Case-II. (e)-(f) Phase portrait of system (\ref{eq:grad}) for Case-III.}
  \label{fig:dynamic}
\end{figure*}

\section{Main Results} \label{sec:theo}

In this section, we present four main results.
We first analyze SAM from a dynamical systems perspective, illustrating its asymptotic behavior and identifying convergence instability at saddle points (\Secref{subsec:dynamic}). We further show that, unlike in conventional optimization methods, saddle points can act as attractors under SAM dynamics (\Secref{subsec:attractor}).

Next, we extend our analysis to stochastic settings. We extend this convergence instability near a saddle point to stochastic system dynamics and further establish theoretical analyses on the difficulty of escaping the saddle point in terms of diffusion (\Secref{subsec:diffusion}). Finally, we demonstrate how training techniques such as momentum and batch size influence this behavior (\Secref{subsec:escape}).

\subsection{Asymptotic Behavior of SAM Dynamics near Saddle Point}
\label{subsec:dynamic}

To investigate the underlying mechanism of SAM, we first apply a qualitative theory of dynamical systems to identify a case of convergence instability under SAM dynamics.
Given the loss function $\ell(\cdot)$, we consider the following gradient flow:
\begin{equation}
\frac{d\boldsymbol{w}}{dt}=-\nabla \ell(\boldsymbol{w}).\label{eq:grad}
\end{equation}
The solution of \eqref{eq:grad} starting from $\boldsymbol{w}_0$ at $t=0$ is called an trajectory, denoted by $\boldsymbol{w}_t$. We call a weight vector $\boldsymbol{w}$ that satisfies $\nabla
\ell(\boldsymbol{w})=0$ {\em equilibrium point} of system (\ref{eq:grad}) and an {\em index-$k$ saddle point} if its Hessian matrix $H_{\ell}(\boldsymbol{w})$ has exactly $k$ negative eigenvalues. The {\em stable manifold} and the {\em unstable manifold} of an index-$k$ saddle point are defined as
\begin{align}
W^s(\boldsymbol{w})&:=\{\boldsymbol{w}_0:
\lim_{t\to\infty}\boldsymbol{w}_t=\boldsymbol{w}\},\\
W^u(\boldsymbol{w})&:=\{\boldsymbol{w}_0:
\lim_{t\to-\infty}\boldsymbol{w}_t=\boldsymbol{w}\}.
\end{align}
where the dimension is $n-k$ and $k$, respectively. Then, the {\em basin of attraction} of a stable (index-zero) equilibrium 
\begin{eqnarray*}
A(\boldsymbol{s}):=\{\boldsymbol{w}_0:
\lim_{t\to\infty}\boldsymbol{w}_t=\boldsymbol{s}\}.
\end{eqnarray*}

Given the two adjacent local minima $\mathbf{s}_1$ and $\mathbf{s}_2$, there exists an index-one equilibrium point $\mathbf{d}$ such that the 1-D unstable manifold $W^u(\mathbf{d})$ converges to both $\mathbf{s}_1$ and $\mathbf{s}_2$ with respect to system (\ref{eq:grad}).
Let us consider the gradient flows near an index-one saddle point $\boldsymbol{d}$ to analyze the behavior of SAM near the basin boundary. 
We illustrate the basin boundaries for two adjacent local minima $\mathbf{s}_1$ and $\mathbf{s}_2$ and their stable and unstable manifolds in \Figref{fig:A1} and \Figref{fig:A1-1}, respectively.
To analyze the behavior of dynamical systems qualitatively, we introduce the Lambda Lemma in our version as follows:
\begin{lemma} \textbf{Lambda Lemma \cite{Guck86, Palis82}.} \label{lemma:lambda}
Let $\vw_t(\Delta) = \{\vw_t : \vw_0 \in \Delta\}$ be the trajectories of the system \eqref{eq:grad}, starting from the points in a 1-D disk $\Delta$ meeting $W^s(\boldsymbol{d})$ transversely. Then, $\vw_t(\Delta)$ approach the unstable manifold $W^u(d)$ as $t \rightarrow \infty$.
\end{lemma}
In other words, the gradient flows near the basin boundary but in $A(\mathbf{s}_1)$ directs to the vector sum of $W^u(\boldsymbol{d})$ and $\mathbf{s}_1$. See Figure \ref{fig:A1} for the illustration (detailed in Appendix).
Thus, the qualitative behavior of the gradient flows starting from $\boldsymbol{w}_t\in A(\boldsymbol{s}_1)$ falls into one of the following cases:
\begin{itemize}
\item\textbf{Case-I.} When $\boldsymbol{w}_t$ is ``away from" an index-one saddle point $\boldsymbol{d}$ and its stable manifold $W^s(\boldsymbol{d})$: $-\nabla \ell(\boldsymbol{w}_t)\sim -\nabla \ell(\boldsymbol{w}_t^p)$ and $\boldsymbol{w}_{t+1}$ converges to $\mathbf{s}_1$ iteratively. (See Figure \ref{fig:A2}).

\item\textbf{Case-II.} When $\boldsymbol{w}_t$ is ``away from" an index-one saddle point $\boldsymbol{d}$ but near $W^s(\boldsymbol{d})$: $-\nabla \ell(\boldsymbol{w}_t)\sim -\nabla \ell(\boldsymbol{w}_t^p)$ and $\boldsymbol{w}_{t+1}$ approaches $W^u(\boldsymbol{d})$ by the Lambda Lemma iteratively. The subsequent weight vector update iteration will fall into the case of (I) or (III). (See Figure \ref{fig:A3}).

\item\textbf{Case-III.} When $\boldsymbol{w}_t$ is near an index-one saddle point $\boldsymbol{d}$ (i.e., $W^s(\boldsymbol{d})\cap B_{\rho}(\boldsymbol{w}_t)\neq \emptyset$): $\boldsymbol{w}_t^p\in A(\mathbf{s}_2)$ outside of $A(\mathbf{s}_1)$ and so $-\nabla\ell(\boldsymbol{w}_t^p)$ directs to $\boldsymbol{s}_2$ and $W^u(\boldsymbol{d})$ by the Lambda Lemma. In this case, there exist two sub-cases.

\begin{enumerate}[label=(\roman*)]
\item When $\boldsymbol{w}_{t+1}\in A(\mathbf{s}_1)$: $\boldsymbol{w}_{t+1}$ approaches $\boldsymbol{d}$ and $W^s(\boldsymbol{d})$ and the subsequent weight vector update iteration will fall into the next case (ii). (See Figure \ref{fig:A4}).
\item  When $\boldsymbol{w}_{t+1}\in A(\mathbf{s}_2)$: starting from $\boldsymbol{w}_{t+1}$, the subsequent weight vector update falls into Case-III but the roles of $\mathbf{s}_1$ and $\mathbf{s}_2$ are reversed. Thus, the gradient oscillates near $W^u(\boldsymbol{d})$. (See Figure \ref{fig:A5}).
\end{enumerate}

\end{itemize}

Therefore, SAM can be hindered by the saddle point $\mathbf{d}$ when the perturbed weight $\vw^p$ falls into a different basin of attraction across a basin boundary (Case-III), whereas GD smoothly directs to the stable equilibrium point.
Furthermore, if $\vw^p$ is consistently located in different basins, $\vw_t$ will not be able to escape the saddle point.

\begin{figure}[t!]
    \centering
    \includegraphics[width=0.68\linewidth]{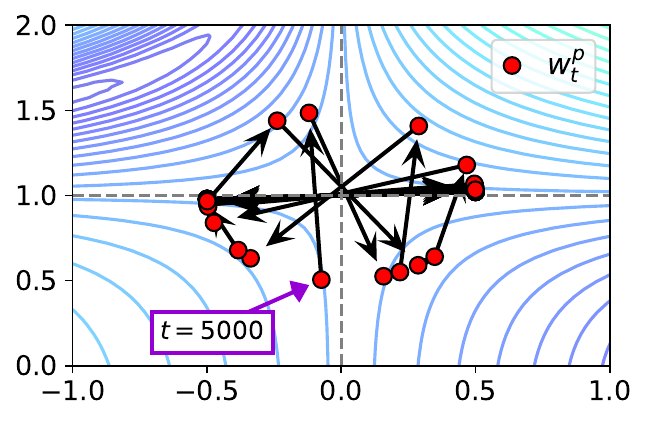} 
    \caption{Trajectory of $\vw^p_t$ after the optimization step $t=5000$, when SAM is beginning to become stuck in the saddle point during SAM optimization in \Figref{fig:intro}. It exhibits exactly the same behavior as that of Case-III in \Figref{fig:dynamic}.}
    \label{fig:wp}
\end{figure}

\begin{figure}[t!]
    \centering
    \includegraphics[width=0.8\linewidth]{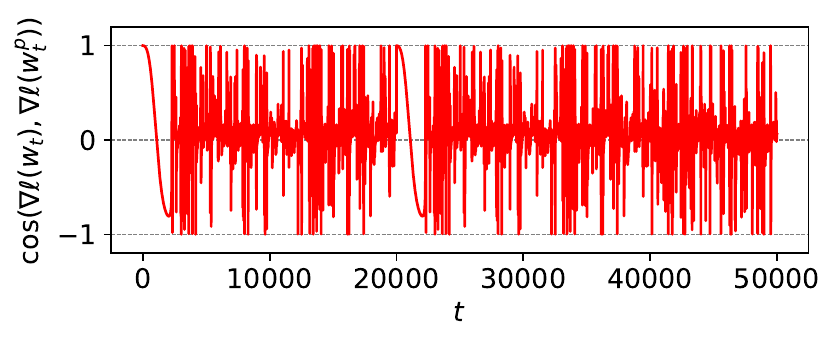} 
    \caption{Gradient oscillation during SAM optimization in \Figref{fig:intro}. The line corresponds to $\cos(\nabla \ell(\vw_t), \nabla \ell(\vw_t^p))$ for optimization step $t$. The oscillation continues until the end of the optimization.}
    \label{fig:toy_grad}
\end{figure}

This analysis can be extended to high-dimensional settings. Under general assumptions on the system of \eqref{eq:grad} \cite{lee2004dynamical}, including the transversality condition, the basin boundary $\partial A(\boldsymbol{s})$ is characterized by the closure of stable manifolds associated with index-one saddle points $\boldsymbol{d}_i$ lying on the boundary, i.e., $\partial A(\boldsymbol{s})=\bigcup_{i} \mathrm{cl}(W^s(\boldsymbol{d}_i)).$
This implies that index-one saddle points $\boldsymbol{d}_i$ behave like attractors near $\partial A(\boldsymbol{s})$.
For more details, we refer the reader to Theorem 3 in \cite{lee2004dynamical} and \cite{jongen1987nonlinear, chiang1996quasi}.

To verify our investigation empirically, we introduce the optimization of the Beale function. It has a single saddle point at $(0, 1)$, which allows us to verify whether the parameter is stuck in the saddle point and the convergence instability near a saddle point. As illustrated in \Figref{fig:dynamic}, it also has four basins with only two basins (top left and bottom right) containing a minimum. We use a learning rate of $\eta$=1e-4, as smaller learning rates often perform better on low dimensional problems.

The optimization result is illustrated in \Figref{fig:intro}. While GD successfully converges to the global minimum, SAM is trapped in the saddle point rather than the global minimum.
To verify the geometric analysis presented in \Figref{fig:dynamic}, the trajectory of the perturbed weight $\vw^p_t$ is plotted in Figure \ref{fig:wp}. After SAM becomes stuck in the saddle point, $\vw^p_t$ continuously crosses the basin boundaries, which is consistent with \Figref{fig:A5}. Additionally, in \Figref{fig:toy_grad}, the cosine between $\nabla \ell (\vw_t)$ and $\nabla \ell (\vw^p_t)$ during the optimization step $t$ is plotted. The cosine value oscillates between $-1$ and $1$, which is consistent with the gradient oscillation described in Case-III-(ii).
In Appendix \ref{ap:double}, we conduct a similar experiment on the double-well potential function $\ell(\theta) = (\mu - \theta^2)^2$. We observe a similar phenomenon that, in several cases, SAM converges to $\theta = 0$, even though the global optima are given by $\theta = \pm \sqrt{\mu}$.
These results demonstrate that SAM becomes stuck in the saddle point as if it were a convergence point.

\begin{figure}
    \centering
  \subfloat[Trajectory of GD and SAM\label{fig:traj}]{%
       \includegraphics[width=0.58\linewidth]{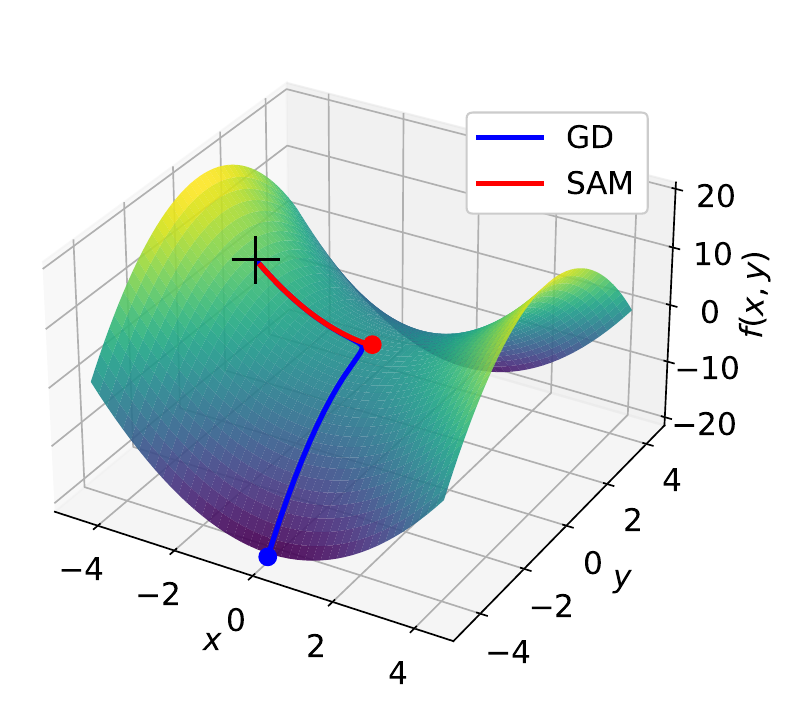}}
    \hfill
  \subfloat[Gradient flow of GD\label{fig:grad_sgd}]{%
        \includegraphics[width=0.38\linewidth]{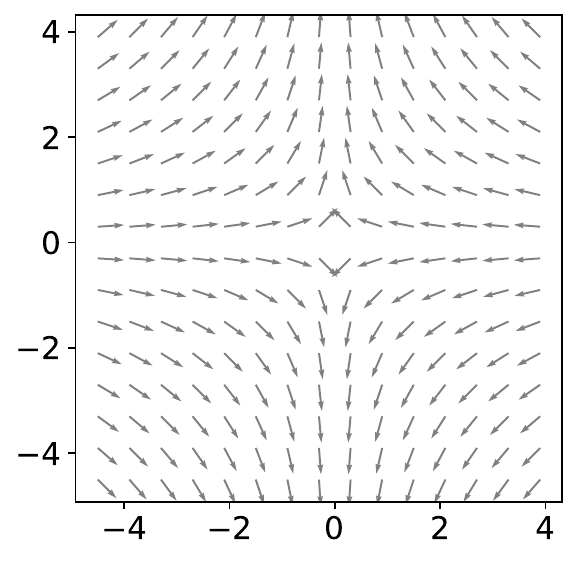}}
  \subfloat[Gradient flow of SAM\label{fig:grad_sam}]{%
        \includegraphics[width=0.38\linewidth]{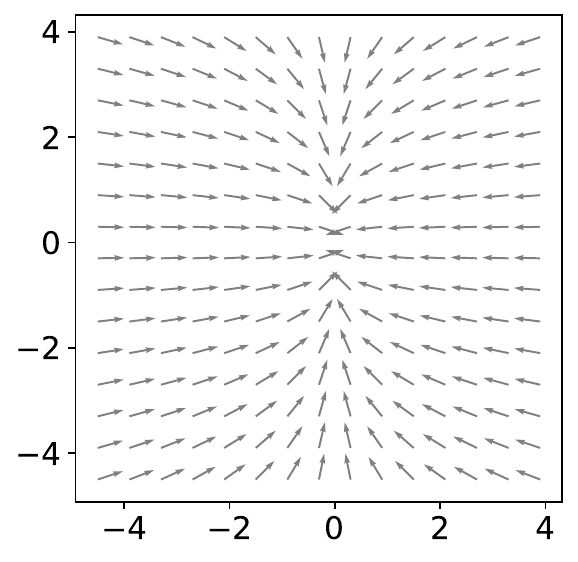}}
  \caption{Optimization on $f(x, y) = x^2 - y^2$ with the saddle point at (0,0). (a) Loss surface with the initial point denoted as the plus symbol $(-3, -\epsilon)$, where $\epsilon=0.01$. (b) Divergence of the gradient flow of GD near the saddle point. (c) Convergence of the gradient flow of SAM with $\rho=1.0$ near the saddle point.}
  \label{fig:toy}
\end{figure}

\subsection{Saddle Point Becomes Attractor in SAM Dynamics} \label{subsec:attractor}
In the previous subsection, we observed that SAM becomes stuck in a saddle point as if the saddle point were a convergence point. Now, we mathematically derive a condition for when a saddle point becomes an attractor under SAM dynamics as follows:
\begin{theorem} \label{thm:attractor}
\label{thm}
Let $\vd$ be an index-one saddle point of system (\ref{eq:grad}) with a negative eigenvalue $\lambda_1$ of the Hessian matrix $H_{\ell}(\vd)$ of the loss function $\ell$.
Then, the saddle point $\vd$ is an attractor of SAM dynamics in \eqref{eq:wp} if $\rho > -1/\lambda_1$.
\end{theorem}
\begin{proof}
See Appendix \ref{ap:proof}.
\end{proof}
Note that the above results can be easily generalized to any type of saddle points provided that
\begin{align}
\lambda_j + {\rho}\lambda_j^2 > 0  \label{eq:lam_cond}
\end{align}
for all the negative eigenvalues $\lambda_j$ at $\vd$. Moreover, this condition can be easily satisfied since the normalized radius $\tilde{\rho}=\rho/\|\nabla\ell(\vw)\|$ in the standard SAM dynamics \cite{foret2020sharpness}, which becomes very large near the saddle point.
This result further suggests that more points can become attractors under SAM dynamic because the term ${\rho} \Lambda^2$ always results in positive diagonal values, where $\Lambda=\mathbf{diag}[\lambda_1, \cdots, \lambda_n]^T$.
Thus, \Thmref{thm:attractor} tells us that a saddle point can become an attractor under SAM dynamics.

\Figref{fig:toy} illustrates the empirical verification of \Thmref{thm:attractor}. We perform an optimization with GD and SAM on a simple function $f(x, y) = x^2 - y^2$, which has a saddle point at $(0, 0)$. In \Figref{fig:traj}, GD (blue-colored) successfully escapes the saddle point due to the benefit of a good initial point $(-3, -\epsilon)$. Specifically, the gradient flow of GD (\Figref{fig:grad_sgd}) demonstrates that the saddle point is not a stable equilibrium point, and thus a slight perturbation at the initial point $\epsilon=0.01$ is sufficient to help GD escape the saddle point.
In contrast, SAM becomes stuck in the saddle point (red-colored in \Figref{fig:traj}), eigenvalues of the Hessian at (0, 0) are $\vlambda=\{2, -2\}$ and thus $\lambda + {\rho} \lambda^2 > 0$ for all eigenvalues $\lambda \in \vlambda$. Therefore, by \Thmref{thm:attractor}, the saddle point becomes an attractor under SAM dynamics as shown in \Figref{fig:grad_sam}.

\begin{figure}
    \centering
  \subfloat[$\lambda_1+{\rho}\lambda_1^2$\label{fig:eigen0}]{%
        \includegraphics[height=0.38\linewidth]{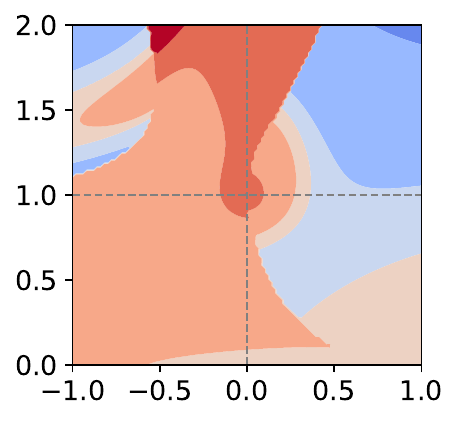}}
  \subfloat[$\lambda_2+{\rho}\lambda_2^2$\label{fig:eigen1}]{%
        \includegraphics[height=0.38\linewidth]{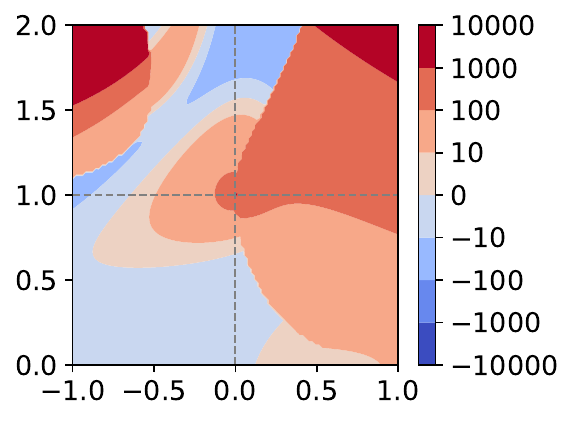}}
  \caption{Value of $\lambda+{\rho}\lambda^2$ for the Beale function in \Figref{fig:intro}. Near the saddle point at (0, 1), for both eigenvalues of the Hessian, $\lambda+{\rho}\lambda^2$ is positive, which indicates that the saddle point becomes an attractor.}
  \label{fig:eigen}
\end{figure}

We further explain the previous optimization result depicted in \Figref{fig:intro} by calculating the eigenvalues of the Hessian for each point in the parameter space.
For each point in the parameter space, we calculate the eigenvalues of the Hessian $\vlambda=\{\lambda_1, \lambda_2\}$ and visualize the heat map of $\lambda + {\rho} \lambda^2$ in \Figref{fig:eigen}. For both $\lambda = \lambda_1$ and $\lambda_2$, $\lambda + {\rho} \lambda^2$ is positive near the saddle point, which is consistent with the behavior of the saddle point as an attractor, as illustrated in \Figref{fig:toy}. This suggests that the convergence instability can occur even for a more complicated loss function because the saddle point may become an attractor under SAM dynamics.

\begin{figure*}
\centering
  \subfloat[Loss landscape\label{fig:nn_intro}]{%
        \includegraphics[height=0.265\linewidth]{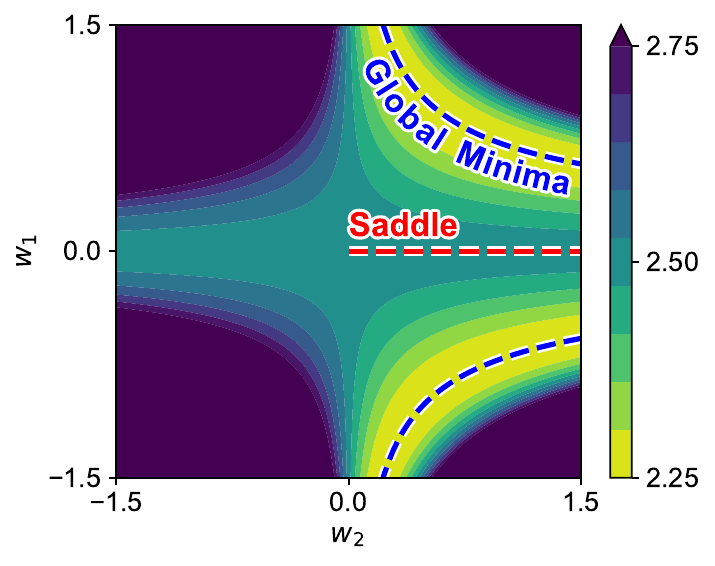}}
  \subfloat[Convergence distribution\label{fig:nn_result_nox}]{%
        \includegraphics[height=0.27\linewidth]{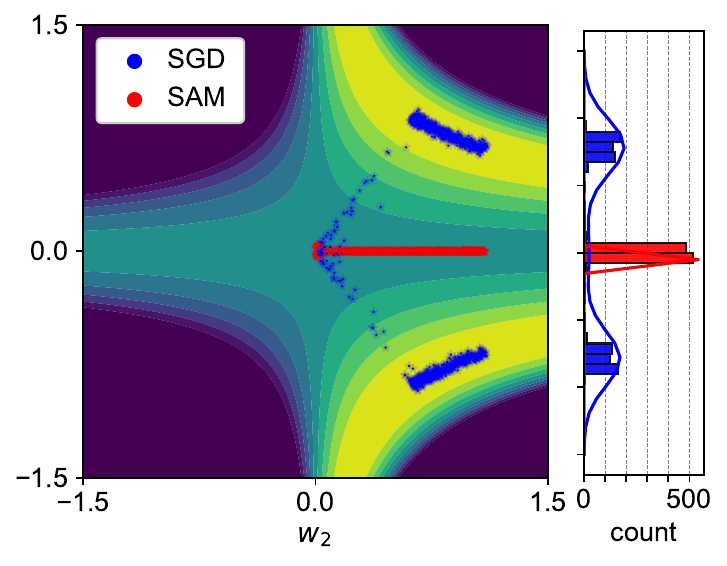}}
  \subfloat[Average loss for different $\rho$\label{fig:nn_result_mean}]{%
        \includegraphics[height=0.265\linewidth]{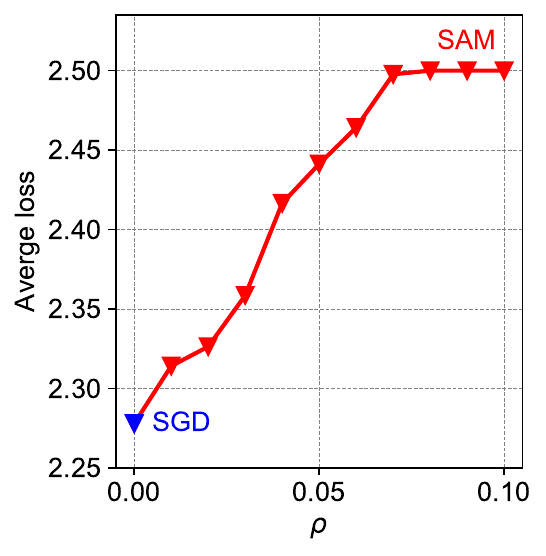}}
  \caption{Toy neural network experiment. (a) Loss landscape for different values of each neuron. (b) Distributions of converged points for SGD and SAM($\rho=0.1$) with the marginal distribution of the parameter $\vw_1$. (c) Average loss of converged points for different $\rho$, where $\rho=0$ indicates SGD.}
  \label{fig:nn}
\end{figure*}

\subsection{Stochastic Behavior of SAM Dynamics with Diffusion near Saddle Point}\label{subsec:diffusion}

In deep learning, mini-batch sampling is a standard practice across various domains \cite{hoffer2017train, ziyin2022sgd}. Accordingly, we analyze the behavior of SAM near saddle points under stochastic system dynamics.
We first establish SAM diffusion with the stochastic differential equation. The basis formulations and notations are borrowed from \cite{risken1996fokker, sato2014approximation, xie2020diffusion, xie2022adaptive}. We assume that the perturbed weight $\vw^p$ is precisely calculated with \eqref{eq:wp}, and thus the stochastic differential equation of SAM dynamics is formalized as follows:
\begin{align}
    d \vw = -\nabla \ell(\vw^p) dt + [\eta C(\vw^p)]^{\frac{1}{2}} dW_t,
    \label{eq:dynamics}
\end{align}
where $\eta$ is a learning rate and $d W_t \sim \mathcal{N}(0, I dt)$ for the identity matrix $I$. $C(\vw)$ is the gradient noise covariance matrix.

Under the dynamics of \eqref{eq:dynamics}, we use the Fokker-Planck equation that describes the probability density of the weight $\vw$ as follows \cite{xie2020diffusion}:
\begin{align}
    \frac{\partial P(\vw, t)}{\partial t} = \nabla\cdot[P(\vw, t)\nabla \ell(\vw^p)] + \nabla\cdot\nabla[D(\vw^p)P(\vw, t)]. \label{eq:fokker}
\end{align}
where $\nabla\cdot$ is the divergence operator and the diffusion matrix is given by $D(\vw^p)={\eta C(\vw^p)}/{2}$.

From now on, to be self-contained, we mainly followed \cite{xie2022adaptive}. Let us denote the loss function of the $i$-th training sample $\ell_i(\vw)$ among $N$ samples in total. By using the Fisher Information Matrix (FIM) \cite{pawitan2001all}, $C(\vw)$ can be approximated near the critical point $\vd$ \cite{jastrzkebski2017three, zhu2018anisotropic}:
\begin{align}\label{eq:C}
    \frac{1}{B}\!\left[\! \frac{1}{N}\!\sum^N_{i=1}\!\nabla\ell_i(\vw) \nabla\ell_i(\vw)^T\!\right]
    \!=\!\frac{1}{B}\text{FIM}(\vw)\!\approx\!\frac{1}{B} [H(\vw)]^+,
\end{align}
where $[A]^+:=V\mathbf{diag}([|\lambda_1|, \cdots, |\lambda_n|]^T)V^T$ for $A=V\mathbf{diag}([\lambda_1, \cdots, \lambda_n]^T)V^T$.
Then, under the second-order Taylor approximation of the loss function $\ell$ near saddle points $\vd$  \cite{mandt2017stochastic, xie2020diffusion},
$D(\vw)$ is independent of $\vw$ near the critical point, i.e.,
\begin{equation}
    D=\frac{\eta}{2B}[H]^+.
    \label{eq:D}
\end{equation}
where $H(\vd)=H$ for simplicity.

Based on these equations, we prove that SAM escapes saddle points more slowly than SGD.

\begin{figure*}
\centering
  \subfloat[(CIFAR-10) Batch size \label{fig:CIFAR10_B}]{%
        \includegraphics[width=0.24\linewidth]{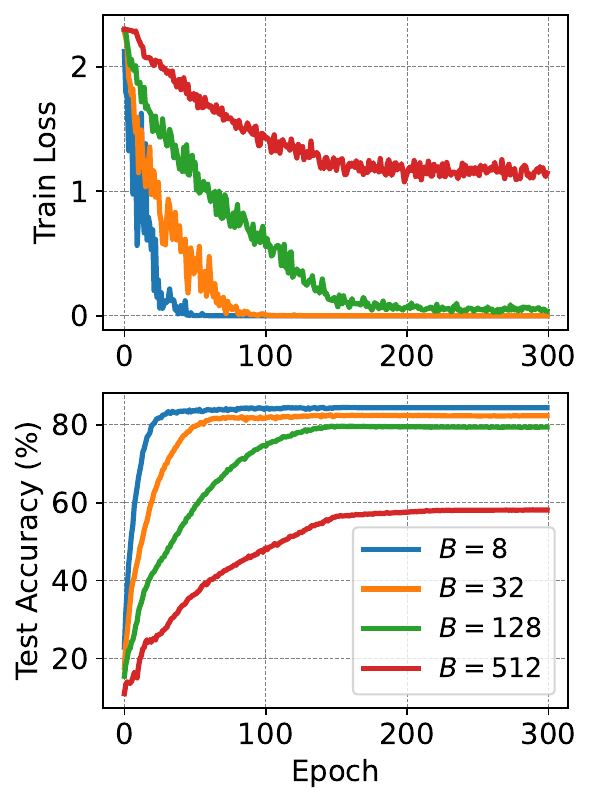}}
  \subfloat[(CIFAR-100) Batch size \label{fig:CIFAR100_B}]{%
        \includegraphics[width=0.24\linewidth]{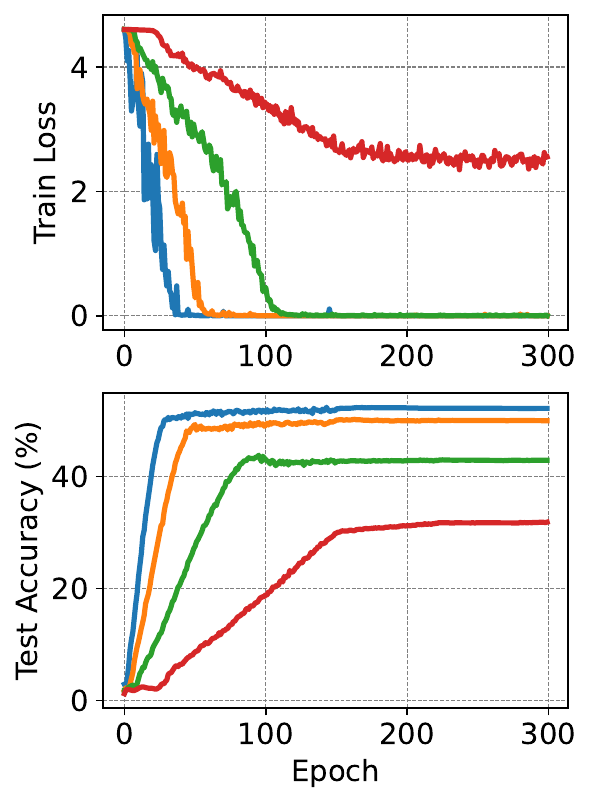}}
  \subfloat[(CIFAR-10) Momentum \label{fig:CIFAR10_M}]{%
        \includegraphics[width=0.24\linewidth]{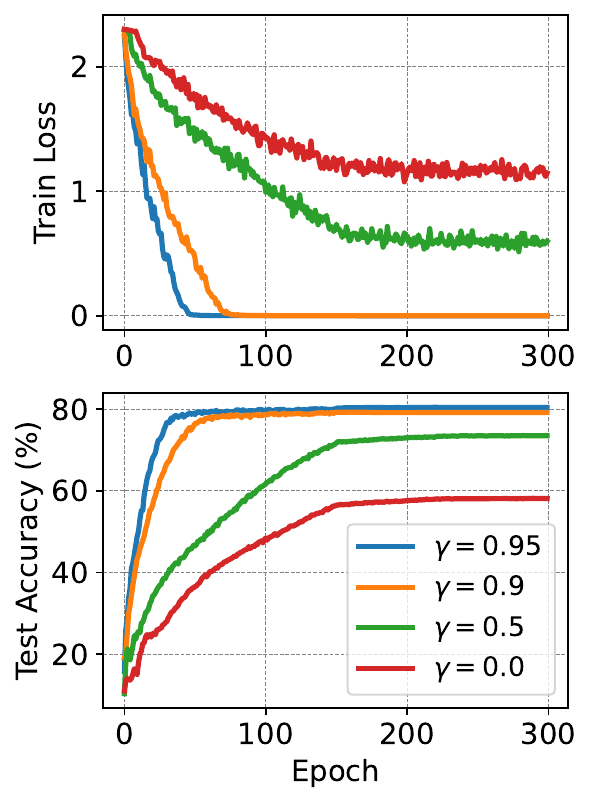}}
  \subfloat[(CIFAR-100) Momentum \label{fig:CIFAR100_M}]{%
        \includegraphics[width=0.24\linewidth]{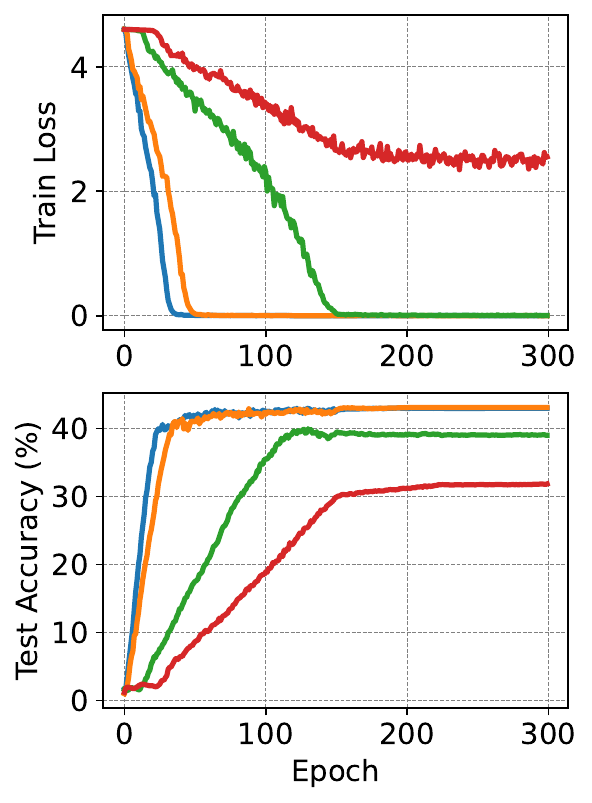}}
  \caption{Effect of varying batch size $B$ and momentum $\gamma$. To measure the pure effect of batch size and momentum, batch normalization and data augmentation are turned off. A smaller batch size and larger momentum lead to better performance of SAM.}
  \label{fig:CIFAR}
\end{figure*}

\begin{theorem}\textbf{(SAM diffusion)}\label{thm:diffusion}
Let us consider a saddle point $\vd=\vw_0$ as the initial parameter under the dynamics of \eqref{eq:dynamics}. Then, under the Fisher Information Matrix approximation \eqref{eq:C} and the second-order Taylor approximation assumption near $\vd$ \cite{xie2022adaptive}, the probability density function of $\vw$ after time $t$ is the Gaussian distribution, i.e., $\vw \sim \mathcal{N}(\vd, Q \mathbf{diag}(\vsigma^2(t)) Q^T)$ with $\vsigma^2(t)=[\sigma^2_1(t), \cdots, \sigma^2_n(t)]^T$, where
\begin{equation}\label{eq:sigma}
 \sigma_j^2 (t) = \frac{\eta|\lambda_j|}{2B\lambda_j(1+{\rho}\lambda_j)^2}[1-\exp(-2\lambda_j(1+{\rho}\lambda_j)^2 t)]
\end{equation}
for the batch size $B$, the column vectors of $Q$ are exactly the eigenvectors of $H(\vd)$, and the $j$-th eigenvalue of $H(\vd)$, $\lambda_j$.
\end{theorem}
\begin{proof} See Appendix \ref{ap:proof}.
\end{proof}
Given that  $\Delta\vw_j(t)=\vw_j(t)-\vw_j(0)$ can be represented by a Gaussian distribution, the mean squared displacement becomes its variance, i.e., $\langle\Delta\vw^2_j(t)\rangle=\sigma^2_j(t)$. 
\begin{corollary}\textbf{(SAM escapes saddle point more slowly than SGD)}\label{cor:diffusion}
Let the mean squared displacement of SAM, denoted by $\Delta_{SAM}:=\langle\Delta\vw^2_j(t)\rangle_{SAM}$, is smaller than that of SGD, i.e., $\Delta_{SGD}:=\langle\Delta\vw^2_j(t)\rangle_{SGD}$, near the saddle point $\vd$. Then, as $|\lambda_j|t \ll 1$  near ill-conditioned saddle points under the condition, the following inequality is satisfied:
\begin{equation}
\Delta_{SGD} - \Delta_{SAM} = \frac{2 \eta t^2 |\lambda_j|^3}{B} {\rho} + \mathcal{O}(B^{-1}\eta t^3 \lambda_j^4 ).
\end{equation}
As $\frac{2 \eta t^2 |\lambda_j|^3}{B} {\rho}$ is always positive, the result implies that SAM escapes saddle points more slowly than SGD.
\end{corollary}
\begin{proof} See Appendix \ref{ap:proof}.
\end{proof}
\Corref{cor:diffusion} tells us that SAM requires more time to escape the saddle point $\vd$ compared to SGD.
Furthermore, $\Delta_{SGD} - \Delta_{SAM}$ increases as ${\rho}$ increases,
which is also consistent with the results from Sections \ref{subsec:dynamic} and \ref{subsec:attractor} that SAM tends to suffer convergence instability as $\rho$ increases.

To validate our theoretical results in stochastic dynamical systems, we conduct an experiment using a neural network as proposed by \citet{ziyin2022sgd}. This experiment involves a two-layer neural network with one neuron, with a non-linear coordinate-wise activation function $\varphi(x)=x^2$. The training set consists of an input $x$ which is fixed to $1$, and $y \in \{-1, 2\}$ with probability of 0.5. The loss function is the mean squared error, and thus the loss landscape yields global minima $\ell(\vw)=2.25$ for $\{(w_1, w_2) | w_1^2=1/{2 w_2}\}$ and saddle points $\ell(\vw)=2.50$ for $\{(w_1, w_2) | w_1 = 0, w_2 \geq 0\}$, as illustrated in \Figref{fig:nn_intro}. 
Considering \Thmref{thm:diffusion}, we initialize the parameters near the saddle point, $w_1$ in the range $[-0.1, 0.1]$ and $w_2$ in the range $[0, 1]$ uniformly. Other settings remain the same as those in \cite{ziyin2022sgd}.

\Figref{fig:nn_result_nox} shows the converged parameter distributions for SGD and SAM over 1,000 random seeds. The results demonstrate that the converged parameters of SAM are mostly saturated in the saddle point area, whereas SGD successfully converges to the global minima.

\Figref{fig:nn_result_mean} shows the average loss of converged points of SAM with varying $\rho$. Here, we normalize the radius with the gradient norm during the optimization for easier comparison with prior studies \cite{foret2020sharpness, zhuang2021surrogate}. The range of $\rho$ in \Figref{fig:nn_result_mean} is commonly used in other benchmark experiments such as the CIFAR classifications \cite{foret2020sharpness, zhuang2021surrogate}. As $\rho$ increases, the average loss of SAM increases.
This result aligns with \Corref{cor:diffusion}, as a larger $\rho$ makes SAM more difficult to escape the saddle point in terms of diffusion.

\subsection{Convergence Instability and Training Tricks}\label{subsec:escape}

The convergence instability of SAM near saddle points, which we observed under both asymptotic and stochastic dynamical systems in previous sections, can lead to suboptimal minimum problems and performance degradation \cite{du2017gradient, kleinberg2018alternative, ziyin2022sgd}. However, this is in contradiction to the claims of prior studies \cite{foret2020sharpness, zhuang2021surrogate}, which state that SAM and its variants generally perform better than other methods on various benchmark datasets. In this section, we theoretically and empirically demonstrate that training tricks, such as momentum and batch size, not only help SAM to escape the saddle point but are also the key to its success.

First, we extend SAM diffusion presented in \Thmref{thm:diffusion} to include momentum and investigate the relationship between the mean squared displacement of SAM, momentum, and batch size.
\begin{theorem}\textbf{(SAM diffusion, momentum, and batch size)}\label{thm:momentum_and_batchsize}
Given a momentum hyper-parameter $\gamma$ and batch size $B$, the mean squared displacement of SAM is given by
\begin{equation}
    \Delta_{SAM} = C_1\frac{(1-e^{-C_2 (1-\gamma)})^2}{(1-\gamma)^3 B} +
    C_3\frac{(1-e^{-\frac{C_4}{1-\gamma}})}{(1-\gamma) B},
    \label{eq:thm3}
\end{equation}
where $C_1=\frac{\eta^2 |\lambda_j|}{2}$, $C_2 = \frac{\eta}{t}$, $C_3=\frac{\eta |\lambda_j|}{2 \lambda_j(1+{\rho}\lambda_j)^2}$, and $C_4 ={2\lambda_j(1+{\rho}\lambda_j)^2 t}$ are positive constants and $\lambda_j$ denotes eigenvalue of the Hessian matrix $H_{\ell}(\vd)$ of loss function $\ell$ at saddle point $\vd$. 
Therefore, $\Delta_{SAM}$ \textbf{increases as (1) momentum $\gamma$ increases and/or (2) batch size $B$ decreases.} Furthermore, under SAM, \textbf{maintaining the same diffusion behavior as in standard SGD requires a larger $\gamma$}, since $\Delta_{\mathrm{SAM}}$ decreases as $\rho$ increases.
\end{theorem}
\begin{proof} See Appendix \ref{ap:proof}.
\end{proof}

\Thmref{thm:momentum_and_batchsize} implies that increasing momentum $\gamma$ and decreasing batch size $B$ can reduce the time to escape a saddle point. This result is consistent with the work of \citet{andriushchenko2022towards}, which observed that a smaller batch size significantly increases the performance of SAM. 

In this subsection, we conduct experiments on the widely used benchmarks CIFAR-10 and CIFAR-100 with ResNet-18. To compare the pure effect of momentum and batch size, we turn off both batch normalization and data augmentation. The default hyperparameters are set to $\rho = 0.1$, $\gamma = 0.0$, and $B = 512$, after which we independently vary $\gamma$ and $B$. The radius is normalized using the gradient norm during training similar to the previous subsection. All models are trained for 300 epochs to ensure convergence.

\Figref{fig:CIFAR} presents the empirical verification of \Thmref{thm:momentum_and_batchsize}.
Figures \ref{fig:CIFAR10_B} and \ref{fig:CIFAR100_B} illustrate that SAM has difficulties in minimizing the training loss as the batch size increases. Notably, for a batch size of $B=512$, SAM shows convergence instability with the train loss exceeding 1, which results in a poor generalization performance of less than 60\% accuracy.
In addition, as shown in Figures \ref{fig:CIFAR10_M} and \ref{fig:CIFAR100_M}, higher values of momentum $\gamma$ lead to better convergence in terms of both training loss and test accuracy.

\begin{figure}[t!]
\centering\includegraphics[width=0.999\linewidth]{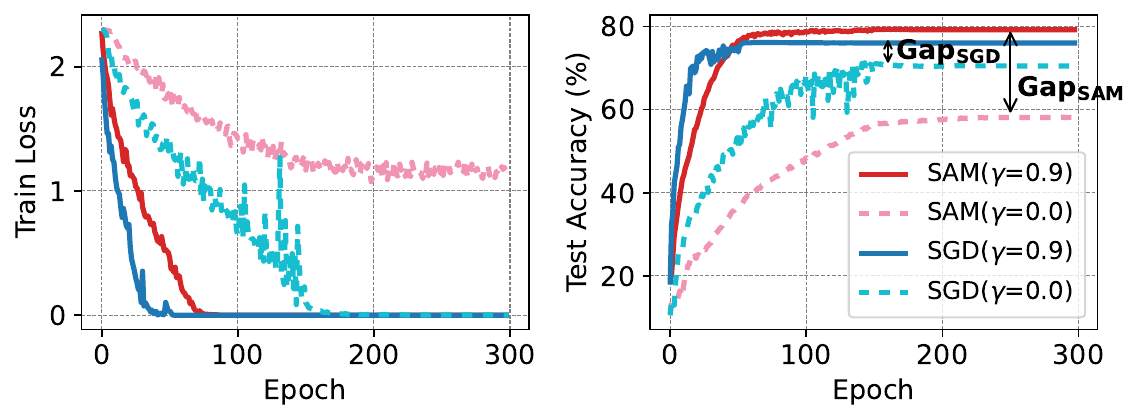} 
    \caption{(CIFAR-10) Effect of momentum $\gamma$ on SGD and SAM. To maintain the same diffusion behavior as in standard SGD, a higher value of $\gamma$ is required as proved in \Thmref{thm:momentum_and_batchsize}. Momentum significantly improves the performance of SAM more dramatically than that of SGD.}
    \label{fig:CIFAR_SGD_SAM}
\end{figure}

\Thmref{thm:momentum_and_batchsize} demonstrates that a higher value of $\gamma$ is required for SAM to maintain the same diffusion behavior as in standard SGD. \Figref{fig:CIFAR_SGD_SAM} shows the train loss and test accuracy for SAM and SGD with $\gamma=0.0$ and $0.9$. The momentum improves the accuracy of SGD by approximately 5\%, while it enhances the accuracy of SAM by more than 20\%.

In \Tabref{tab:momentum}, we evaluate the test accuracy of SAM for different radius $\rho$ and momentum $\gamma$. To achieve optimal performance, we enable batch normalization and apply data augmentation. The reported values represent the average and standard deviation of the test accuracy over three different random seeds. The minimum accuracies are observed only for $\gamma=0.0$, whereas the best performance is observed for $\gamma=0.9$ or $0.95$. Furthermore, as $\rho$ increases, the gap between the best and worst performances increases as proved in \Thmref{thm:momentum_and_batchsize}. Considering the fact that $\rho=0.1$ with $\gamma=0.95$ yields the best accuracy over all combinations, using a higher momentum $\gamma$ with a larger $\rho$ can be beneficial to improve the generalization performance.

\begin{table}[t!]
\centering
\caption{(CIFAR-10) Test acc. for different momentum $\gamma$ and radius $\rho$. The \textbf{bold} and \underline{underline} denote the maximum and minimum accuracy for each $\rho$, respectively. Batch normalization and data augmentation are turned on.}
\label{tab:momentum}
\renewcommand\arraystretch{1.1}

\resizebox{\columnwidth}{!}{%
\begin{tabular}{c|rrr|c}  \hline
SAM & \multicolumn{3}{c|}{Momentum $\gamma$}  & Gap          \\ 
Radius $\rho$ & \multicolumn{1}{c}{0.0} & \multicolumn{1}{c}{0.9}           & \multicolumn{1}{c|}{0.95} & (Max - Min)          \\ \hline
0.01  & \underline{93.77$\pm$0.08}
        & \textbf{94.99$\pm$0.09}
            & 94.94$\pm$0.09 & 1.22
          \\
0.05  & \underline{93.56$\pm$0.03}
        & \textbf{95.01$\pm$0.11}
            & 94.94$\pm$0.11 & 1.45
          \\
0.1   & \underline{92.29$\pm$0.08}
        & 94.84$\pm$0.15
          & \textbf{95.08$\pm$0.07} & 2.80
 \\
0.5   & \underline{86.20$\pm$1.66}
        & 88.76$\pm$1.14
          & \textbf{91.98$\pm$0.03} & 5.79
 \\ \hline
\end{tabular}%
}
\end{table}

\section{Limitations and Future Work}

Our analysis of the behavior of SAM through dynamical systems reveals its difficulty in converging near saddle points. However, it is important to note that SAM generally improves performance \cite{foret2020sharpness, chen2021vision}. This observation is also supported by several theoretical studies, particularly in terms of error bounds and others \cite{behdin2023statistical, bartlett2023dynamics}. Therefore, further work is needed to integrate these findings in more general settings and to better understand the behavior of SAM in the context of complex neural networks.

Moreover, our results highlight that momentum and batch size, which are typically fixed in practice, could also be explored under SAM dynamics. Although this direction involves complex challenges, exploring these aspects could contribute to the development of new optimization methods \cite{tan2025stabilizing, li2024implicit}. Additionally, \citet{chen2023stochastic} argued that the generalization benefit of SAM may stem from transient attraction to saddle points. However, our findings suggest that permanent trapping in saddle regions can become a failure mode. Taken together, the roles of momentum and batch size can be further investigated in terms of balancing saddle exploration and escape dynamics.

\section{Conclusion}
To gain insight into SAM dynamics, we analyzed its convergence instability near a saddle point. Through theoretical and empirical analyses, we demonstrated that this convergence problem can occur in various tasks, from a simple optimization to complicated neural network training. Additionally, we argued the importance of momentum and batch size in relation to the diffusion theory. We believe that our work sheds light on understanding the mechanism of SAM and hope that future work will build upon our findings to develop new training methods.

\section*{Impact Statement}
While advancements in optimization algorithms broadly contribute to the capabilities of deep learning models, this work is theoretical and methodological in nature. Therefore, there are no specific societal consequences or ethical aspects of this work.

\section*{Acknowledgement}
This work was supported in part by multiple grants from the Korean government (MSIT).
Hoki Kim was supported by the National Research Foundation of Korea (NRF) grant funded by the Korea government (MSIT) (RS-2026-25484948) (Contribution: 30\%), and by the IITP (Institute of Information \& Communications Technology Planning \& Evaluation)-ITRC (Information Technology Research Center) grant funded by the Korea government (MSIT) (IITP-2026-RS-2024-00438056) (Contribution: 20\%).
Yujin Choi was supported by the National Research Foundation of Korea (NRF) grant funded by the Korea government (MSIT) (RS-2025-00515481) (Contribution: 10\%).
Jinseong Park was supported by a KIAS Individual Grant (AP102301, AP102303) through the Center for AI and Natural Sciences at Korea Institute for Advanced Study  (Contribution: 10\%).
Jaewook Lee was partly supported by the Institute of Information \& Communications Technology Planning \& Evaluation (IITP) grant funded by the Korea government (MSIT) (No. RS-2022-II220984, Development of Artificial Intelligence Technology for Personalized Plug-and-Play Explanation and Verification of Explanation) (Contribution: 20\%), and by the National Research Foundation of Korea (NRF) grant funded by the Korean government (MSIT) (No. RS-2024-00338859) (Contribution: 10\%).


\bibliography{example_paper}
\bibliographystyle{icml2026}

\newpage
\appendix
\onecolumn

\section{Proofs}\label{ap:proof}
Without loss of generality, we assume that (\ref{eq:grad}) is {\em hyperbolic} so that the Hessian matrix of $\ell$ at each equilibrium point has no zero eigenvalues, which is a generic property that holds for typical loss functions. Then, the Hessian matrix has real eigenvalues because it is a real symmetric matrix.

\subsubsection*{Proof of \Thmref{thm:attractor}}
\begin{proof}
Given the gradient flow of SAM in \eqref{eq:wp}, we have
\begin{align}
\mathbf{F}(\vw) &= - \nabla\ell(\vw + {\rho} \nabla \ell(\vw)) \\
D\mathbf{F}(\vw)  &= - \nabla^2\ell(\vw + {\rho} \nabla \ell(\vw))[I+ {\rho} \nabla^2 \ell(\vw)],
\label{eq:hessian_adaprho}
\end{align}
where $D\mathbf{F}$ is the Jacobian matrix of $\mathbf{F}$. We note that the notations are aligned with \cite{perko2013differential}.
Since the Hessian matrix is a real symmetric matrix, $H_{\ell}(\vd)=Q\Lambda Q^T$, where $Q$ is a real orthogonal matrix and $\Lambda=\mathbf{diag}[\lambda_1, \cdots, \lambda_n]^T$ is a real diagonal matrix consisting of eigenvalues as its elements. Given a saddle point $\vd$, we have $\nabla \ell(\vd)=0$ and thus \eqref{eq:hessian_adaprho} becomes
\begin{align}
D\mathbf{F}(\vw)|_{\vw=\vd} &= - \nabla^2\ell(\vd + {\rho} \nabla \ell(\vd))[I+ {\rho} \nabla^2 \ell(\vd)] \\
&= - Q(\Lambda + {\rho} \Lambda^2)Q^T.
\end{align}
As all the diagonal elements of $\Lambda + {\rho} \Lambda^2$ are positive, $\vd$ becomes an attractor of SAM dynamics.
\end{proof}

\subsubsection*{Proof of \Thmref{thm:diffusion}}
\begin{proof}
The solution of the Fokker-Planck Equation \eqref{eq:fokker} should be formalized as follows:
\begin{equation}
    P(\vw, t) = (\prod_{j=1}^{n}2\pi\sigma_j)^{-\frac{1}{2}}\exp\left(-\frac{1}{2} (\vw - \vd)^T Q \mathbf{diag}(\vsigma^2(t)) Q^T (\vw - \vd)\right).
\end{equation}

To be self-contained, we mainly followed Appendix A.1 in \cite{xie2022adaptive}. Let us denote the loss function of the $i$-th training sample $\ell_i(\vw)$ among $N$ samples in total. By using the Fisher Information Matrix (FIM) \cite{pawitan2001all}, the following approximation can be adopted near the critical point $\vd$ as described in \cite{jastrzkebski2017three, zhu2018anisotropic, xie2022adaptive}.
\begin{align}
    C(\vw) \approx \frac{1}{B} \left[ \frac{1}{N} \sum^N_{i=1} \nabla\ell_i(\vw) \nabla\ell_i(\vw)^T  \right] = \frac{1}{B}\text{FIM}(\vw) \approx \frac{1}{B}[H(\vw)]^+.
\end{align}
Here, given $A=V\mathbf{diag}([\lambda_1, \cdots, \lambda_n]^T)V^T$, we denote $V\mathbf{diag}([|\lambda_1|, \cdots, |\lambda_n|]^T)V^T$ as $[A]^+$.
By the above equation and second-order Taylor approximation, near the critical point, $D(\vw)$ is independent of $\vw$, i.e.,
\begin{equation}
    D=\frac{\eta}{2B}[H]^+.
\end{equation}
where $H(\vd)=H$ for simplicity.

Following \cite{xie2022adaptive}, without loss of generality, we consider one-dimensional solution. Under the second-order Taylor approximation assumption, we have
\begin{align*}
    \nabla\ell(\vw^p) &= \nabla \left[\ell(\vd) + \frac{1}{2} (\vw^p - \vd)^T H (\vw^p - \vd) \right] \\
    &= H[\vw+{\rho}\nabla\ell(\vw)]\nabla[\vw+{\rho}\nabla\ell(\vw)] = H[I+{\rho}H]^2\vw.
\end{align*}
Then, the right term of \eqref{eq:fokker} can be formalized as
\begin{align*}
    &\nabla\cdot[P(\vw, t)\nabla \ell(\vw^p)] + \nabla\cdot\nabla[D(\vw^p)P(\vw, t)] \\&= P(\vw, t)H[I+{\rho}H]^2 - H[I+{\rho}H]^2 \cdot \frac{\vw}{\sigma^2} P(\vw, t) +D\left(\frac{\vw^2}{\sigma^4} - \frac{1}{\sigma^2}\right) P(\vw,t) \\
    &= \left(1 - \frac{\vw^2}{\sigma^2} \right)H[I+{\rho}H]^2P(\vw,t)  + D\left(\frac{\vw^2}{\sigma^4} - \frac{1}{\sigma^2}\right)P(\vw,t) \\
    &= \left( - \sigma^2 H[I+{\rho}H]^2 + D \right)\left(\frac{\vw^2}{\sigma^4} - \frac{1}{\sigma^2}\right)P(\vw,t).
\end{align*}
On the other hand, the left term of \eqref{eq:fokker} can be formalized as
\begin{align*}
    \frac{\partial P(\vw,t)}{\partial t} = \frac{1}{2} \left(\frac{\vw^2}{\sigma^4} - \frac{1}{\sigma^2}\right) P(\vw, t) \frac{\partial \sigma^2}{\partial t}.
\end{align*}
Thus, the solution of \eqref{eq:fokker} is
\begin{align*}
    \frac{\partial \sigma^2}{\partial t} = 2D - 2 \sigma^2 H[I+{\rho}H]^2,
\end{align*}
By using the initial condition $\sigma^2(0)=0$ and \eqref{eq:D}, we can obtain
\begin{align*}
    \sigma_j^2(t) = \frac{\eta|\lambda_j|}{2B\lambda_j(1+{\rho}\lambda_j)^2}[1-\exp(-2\lambda_j(1+{\rho}\lambda_j)^2 t)].
\end{align*}
\end{proof}

\subsubsection*{Proof of \Corref{cor:diffusion}}

\begin{proof}
By the definition of the mean squared displacement, $\langle\Delta\vw^2_j(t)\rangle$ is equal to $\sigma_j^2 (t)$. Thus, given \Thmref{thm:diffusion}, we have
\begin{align*}
\Delta_{SGD} &= \frac{\eta|\lambda_j|}{2B\lambda_j}[1-\exp(-2\lambda_j t)] \\
\Delta_{SAM} &= \frac{\eta|\lambda_j|}{2B\lambda_j(1+{\rho}\lambda_j)^2}[1-\exp(-2\lambda_j(1+{\rho}\lambda_j)^2 t)].
\end{align*}
As $|\lambda_j|t \ll 1$ near ill-conditioned saddle points \cite{xie2022adaptive},  the difference $\Delta_{SGD} - \Delta_{SAM}$ can be formalized as follows:
\begin{align*}
\Delta_{SGD} - \Delta_{SAM} &=  \frac{\eta|\lambda_j|}{2B\lambda_j(1+{\rho}\lambda_j)^2 }\left[(1+{\rho}\lambda_j)^2[2\lambda_j t - 2(\lambda_j t)^2] - [2\lambda_j(1+{\rho}\lambda_j)^2 t -2(\lambda_j(1+{\rho}\lambda_j)^2 t)^2]\right] + \mathcal{O}(B^{-1}\eta t^3 \lambda_j^4 )
\\& = \frac{2\eta t^2|\lambda_j|^3  {\rho}  }{B} + \mathcal{O}(B^{-1}\eta t^3 \lambda_j^4 ).
\end{align*}
The first equation is derived by the Taylor expansion of $\exp$, and the fact that $|\lambda_j|t \ll 1$ near ill-conditioned saddle points is used for both equations.

\end{proof}

\subsubsection*{Proof of \Thmref{thm:momentum_and_batchsize}} \label{ap:momentum_and_batchsize}

\begin{proof}
Similar to \Thmref{thm:diffusion}, we consider the on-dimensional case near a critical point. As the momentum dynamics with a momentum $\gamma$ and damping $\tau$ can be written as follows:
\begin{align*}
    \vm_t &= \gamma \vm_{t-1} + (1-\tau) \vg_t \\
    \vw_{t+1} &= \vw_t - \eta \vm_t
\end{align*}
We let $dt=\eta$, $\phi=\frac{1-\gamma}{dt}$, and $M = \frac{dt}{1-\tau}$.
Then, the stochastic differential equation and its Fokker-Planck Equation becomes
\begin{align*}
    M d\dot{\vw}&= - \phi d\vw-\frac{\partial \ell(\vw)}{\partial \vw} dt + [2D]^{1/2}dW_t \\
    \frac{\partial P(\vw, \vb, t)}{\partial t} &= - \nabla_\vw \cdot [\vb P(\vw, \vb, t)] + \nabla_\vb [\phi\vm + M^{-1} \nabla_\vw \ell (\vw)]P(\vw, \vb, t) + \nabla_\vb \cdot M^{-2} D \cdot \nabla_{\vb} P(\vw, \vb, t).
\end{align*}
where $\dot{\vw}:=\frac{d\vw}{dt}$. Further, combining the proof of Theorem 2 in \cite{xie2022adaptive} and \Thmref{thm:diffusion} in the main paper, we have the mean squared displacement of SAM with $\rho$ for $\tau=0$,
\begin{align*}
    \langle\Delta\vw^2_j(t)\rangle &=
    \frac{\eta |\lambda_j|}{2\phi^3 M^2 B}[1-\exp(-\phi t)]^2 +
    \frac{\eta|\lambda_j|}{2\phi M B\lambda_j(1+{\rho}\lambda_j)^2}\left[1-\exp\left(-\frac{2\lambda_j(1+{\rho}\lambda_j)^2 t}{\phi M}\right)\right] \\
    &=
    \frac{\eta^2 |\lambda_j|}{2 (1-\gamma)^3  B}\left[1-\exp\left(-\frac{1-\gamma}{\eta}t\right)\right]^2 +
    \frac{\eta |\lambda_j|}{2 (1-\gamma) B\lambda_j(1+{\rho}\lambda_j)^2}\left[1-\exp\left(-\frac{2\lambda_j(1+{\rho}\lambda_j)^2 t}{1-\gamma}\right)\right].
\end{align*}
Let $C_1=\frac{\eta^2 |\lambda_j|}{2}$, $C_2 = \frac{\eta}{t}$, $C_3=\frac{\eta |\lambda_j|}{2 \lambda_j(1+{\rho}\lambda_j)^2}$, and $C_4 ={2\lambda_j(1+{\rho}\lambda_j)^2 t}$. Note that $C_1, C_2, C_3$, and $C_4$ are positive. The above equation can be reformulated as follows:
\begin{align*}
    \langle\Delta\vw^2_j(t)\rangle = \frac{C_1}{(1-\gamma)^3 B}\left[1-\exp\left(-C_2 (1-\gamma)\right)\right]^2 +
    \frac{C_3}{(1-\gamma) B}\left[1-\exp\left(-\frac{C_4}{1-\gamma}\right)\right].
\end{align*}
The second term in the right-hand side is an increasing function with respect to $\gamma\in[0,1]$, since both $\frac{C_3}{(1-\gamma)}$ and $1-\exp\left(-\frac{C_4}{1-\gamma}\right)$ are increasing functions with respect to $\gamma$. For the first term in the right-hand, we use the following function.
\begin{align*}
    h(\gamma) = \frac{1}{(1-\gamma)^3}\left[1-e^{-C_2(1-\gamma)}\right]^2.
\end{align*}
We have $h(0)=(1-e^{-C_2})^2>0$ and
\begin{align*}
    h'(\gamma) = \frac{e^{-2C_2(1-\gamma)}[e^{C_2(1-\gamma)}-1]}{(1-\gamma)^4}\left[3 (e^{C_2(1-\gamma)} - 1) - 2 C_2 (1-\gamma)  \right] > 0,
\end{align*}
since $e^{C_2(1-\gamma)}-1 \geq C_2(1-\gamma)$ for $\gamma \in [0,1]$ and $C_2 > 0$. Therefore, $h(\gamma)$ is an increasing function with respect to $\gamma\in[0,1]$. Thus, $\langle\Delta\vw^2_j(t)\rangle$ is also an increasing function with respect to $\gamma$. 

Furthermore, consider the effect of $\rho$ on the diffusion behavior. The partial derivative of the mean squared displacement with respect to $\rho$ is given by
\begin{align*}
\frac{\partial \langle\Delta\vw^2_j(t)\rangle}{\partial \rho}
= \frac{\partial}{\partial \rho} \left[
\frac{\eta |\lambda_j|}{2 (1 - \gamma) B \lambda_j (1 + \rho \lambda_j)^2}
\left( 1 - \exp\left( -\frac{2 \lambda_j (1 + \rho \lambda_j)^2 t}{1 - \gamma} \right) \right)
\right].
\end{align*}
Define $f(\rho) := \lambda_j(1 + \rho \lambda_j)^2$ and $g(x) := \frac{k_1}{x}\left[1 - \exp(-k_2 x)\right]$ for constants $k_1, k_2 > 0$. Since $\frac{\partial f(\rho)}{\partial \rho} = 2 \lambda_j^2 (1 + \rho \lambda_j)$ and $|\lambda_j| t \ll 1$ near saddle points, $f(\rho)$ is strictly increasing in $\rho$. Also, we have
\begin{align*}
\frac{\partial g(x)}{\partial x} =\frac{k_1}{x^2}[(1+k_2 x)\exp(-k_2 x)-1].
\end{align*}
Since the inequality $\exp(x) \geq 1 + x$, we have $\frac{\partial g(x)}{\partial x} \leq 0$ for all $x \neq 0$, implying that $g(x)$ is strictly decreasing on both $(-\infty, 0)$ and $(0, \infty)$. Since $g(x)$ is continuous at $x = 0$, the composite function $g(f(\rho))$ decreases as $\rho$ increases. Therefore, to maintain the same diffusion behavior as in standard SGD (i.e., $\rho=0$) requires a larger $\gamma$ under SAM dynamics with $\rho>0$.
\end{proof}

\section{Additional Experiment on Double-Well Potential}\label{ap:double}
In addition to the Beale function, we evaluate the behavior of SAM on the double-well potential function, defined as $\ell(\theta) = (\mu - \theta^2)^2$. 
Our experimental setup is as follows: $\mu=1$, learning rate $\eta=0.01$, total iterations $T=2000$, batch size $N=8000$, and initialization $\theta_0 \sim U[-2.5\sqrt{\mu}, 2.5\sqrt{\mu}]$. We empirically verified that $T=2000$ iterations are sufficient to observe consistent asymptotic behavior.

The global minima of double-well potential function $\ell(\theta) = (\mu - \theta^2)^2$ is at $\theta = \pm\sqrt{\mu}$. The SAM objective $\max_{|\varepsilon| \leq \rho} \ell(\theta + \varepsilon)$ has a phase transition at a critical perturbation radius of $\rho_c = \sqrt{\mu/2}$.

\begin{table}[H]
\centering
\caption{Probability of convergence to the global optimum and saddle trap for varying $\rho$.}
\begin{tabular}{rrr}
\toprule
$\rho$ & $P(\text{global opt})$ & $P(\text{saddle trap})$ \\
\midrule
0.000 & 1.0000 & 0.0000 \\
0.085 & 0.1912 & 0.0311 \\
0.170 & 0.0000 & 0.0670 \\
0.255 & 0.0000 & 0.0973 \\
\bottomrule
\end{tabular}
\begin{tabular}{rrr}
\toprule
$\rho$ & $P(\text{global opt})$ & $P(\text{saddle trap})$ \\
\midrule
0.340 & 0.0000 & 0.1379 \\
0.425 & 0.0000 & 0.1716 \\
0.595 & 0.0000 & 0.2410 \\
0.680 & 0.0000 & 0.2765 \\
\bottomrule
\end{tabular}
\label{tab:saddle_trap_prob}
\end{table}

In \Tabref{tab:saddle_trap_prob}, we estimate $P(\text{saddle trap})$, defined as the fraction of trials where the final iterate satisfies $|\theta_T| \leq 0.03\sqrt{\mu}$. 
For $\rho < \rho_c$, the central region ($\theta \approx 0$) remains suboptimal, meaning that near-zero convergence constitutes a genuine saddle trap. For $\rho \geq \rho_c$, the center is no longer suboptimal and, for sufficiently large $\rho$, becomes globally optimal. As shown in our results, $P(\text{saddle trap})$ is positive below $\rho_c$ but drops to zero at and above this threshold. This demonstrates that SAM is prone to getting stuck at saddle points rather than converging to global minima. This phenomenon consistent to the case of the Beale function. We believe this observation provides strong empirical support for the theoretical claims established in \Thmref{thm:attractor}.


\end{document}

%% file: math_commands.tex
\usepackage{amsmath,amsfonts,bm}

\def\Figref#1{Figure~\ref{#1}}
\def\Tabref#1{Table~\ref{#1}}
\def\Thmref#1{Theorem~\ref{#1}}

\def\Corref#1{Corollary~\ref{#1}}

\def\Secref#1{Section~\ref{#1}}

\def\eqref#1{equation~\ref{#1}}

\def\1{\bm{1}}

\def\0{{\bm{0}}}
\def\1{{\bm{1}}}

\def\vlambda{\bm{\lambda}}

\def\vsigma{\bm{\sigma}}

\def\vb{{\bm{b}}}

\def\vd{{\bm{d}}}

\def\vg{{\bm{g}}}

\def\vm{{\bm{m}}}

\def\vv{{\bm{v}}}
\def\vw{{\bm{w}}}
\def\vx{{\bm{x}}}
\def\vy{{\bm{y}}}

\DeclareMathAlphabet{\mathsfit}{\encodingdefault}{\sfdefault}{m}{sl}
\SetMathAlphabet{\mathsfit}{bold}{\encodingdefault}{\sfdefault}{bx}{n}



%% file: my_math_commands.tex
\usepackage{color,soul}
\usepackage{enumitem}
\usepackage{nicefrac}
\usepackage{booktabs}

\renewcommand{\eqref}[1]{(\ref{#1})}

\newtheorem{theorem}{Theorem}
\newtheorem*{theorem*}{Theorem}
\newtheorem{corollary}{Corollary}
\newtheorem{lemma}{Lemma} 
\newtheorem*{lemma*}{Lemma}
\def\0{\bm{0}}